\documentclass[preprint,5p,twocolumn]{elsarticle}

\usepackage{amssymb}
\usepackage{amsmath}
\usepackage{amsfonts}

\usepackage{multirow}
\usepackage{siunitx}
\usepackage{makecell}
\usepackage{booktabs}
\usepackage{subcaption}
\usepackage{hyperref}

\journal{Information Fusion}

\begin{document}

\begin{frontmatter}

\title{Distilling Implicit Multimodal Knowledge into Large Language Models for Zero-Resource Dialogue Generation}

\author[label1]{Bo Zhang}
\ead{zhangbo1998@mail.dlut.edu.cn}

\author[label2]{Hui Ma}
\ead{huima@hfut.edu.cn}

\author[label1]{Jian Ding}
\ead{91mr_ding@mail.dlut.edu.cn}

\author[label1]{Jian Wang \corref{cor1}}
\ead{wangjian@dlut.edu.cn}

\author[label1]{Bo Xu}
\ead{xubo@dlut.edu.cn}

\author[label1]{Hongfei Lin}
\ead{hflin@dlut.edu.cn}

\cortext[cor1]{Corresponding author}

%\address{School of Computer Science and Technology, Dalian University of Technology, Dalian, 116024, Liaoning, China}
\affiliation[label1]{organization={School of Computer Science and Technology, Dalian University of Technology},%Department and Organization
            city={Dalian},
            postcode={116024}, 
            state={Liaoning},
            country={China}}

\affiliation[label2]{organization={School of Computer Science and Information Engineering, Hefei University of Technology},%Department and Organization
            city={Hefei},
            postcode={230601}, 
            state={Anhui},
            country={China}}

\begin{abstract}
Integrating multimodal knowledge into large language models (LLMs) represents a significant advancement in dialogue generation capabilities. However, the effective incorporation of such knowledge in zero-resource scenarios remains a substantial challenge due to the scarcity of diverse, high-quality dialogue datasets. To address this, we propose the Visual Implicit Knowledge Distillation Framework (VIKDF), an innovative approach aimed at enhancing LLMs for enriched dialogue generation in zero-resource contexts by leveraging implicit multimodal knowledge. VIKDF comprises two main stages: knowledge distillation, using an Implicit Query Transformer to extract and encode visual implicit knowledge from image-text pairs into knowledge vectors; and knowledge integration, employing a novel Bidirectional Variational Information Fusion technique to seamlessly integrate these distilled vectors into LLMs. This enables the LLMs to generate dialogues that are not only coherent and engaging but also exhibit a deep understanding of the context through implicit multimodal cues, effectively overcoming the limitations of zero-resource scenarios. Our extensive experimentation across two dialogue datasets shows that VIKDF outperforms existing state-of-the-art models in generating high-quality dialogues. The code is available at \url{https://github.com/zhangbo-nlp/VIKDF}.
\end{abstract}


\end{frontmatter}


\section{Introduction}
Dialogue generation, a pivotal component of natural language processing, aims to create responses that are both natural and engaging within specific dialogue contexts. The emergence of large language models (LLMs), such as the Generative Pre-trained Transformer (GPT) series \cite{ref1, ref2, ref3}, has marked significant advancements in this domain. These models excel in identifying complex linguistic patterns and semantic details due to their training on extensive textual datasets. However, their effectiveness is limited to text-based contexts, overlooking the rich, multimodal aspects of human dialogue that incorporate visual, auditory, and other sensory inputs. This limitation highlights a crucial challenge: enabling LLMs to navigate the multimodal nature of human interactions, a capability that humans possess inherently.

The integration of multimodal knowledge into dialogue systems signifies a major progression towards more nuanced and human-like communication capabilities. It enables these systems to understand and interpret the nuances of human communication that transcend beyond mere text, capturing the essence of multimodal interactions \cite{ref4, ref5, ref6, ref48}. Building upon this concept, there has been an increase in research focused on augmenting LLMs with multimodal knowledge \cite{ref7, ref8, ref9, ref10}. This involves processing and understanding information across different modalities, such as images, videos, and audio, thereby equipping LLMs to perform tasks that necessitate cross-modal comprehension. While these developments significantly expand the capabilities of LLMs in engaging with multimodal content, challenges persist in effectively applying multimodal knowledge in dialogue generation, necessitating further exploration and innovation in this area.

One pivotal challenge in augmenting LLMs with multimodal capabilities, crucial for advancing human-like dialogue generation, is the scarcity of high-quality, diverse multimodal dialogue datasets. This is particularly notable in domains that demand intricate interactions, such as image-grounded dialogues \cite{ref11}. Image-grounded dialogues involve conversations anchored on a shared image, necessitating visual reasoning and a wealth of common sense to elicit coherent and engaging responses. Current datasets \cite{ref11, ref12, ref13}, while foundational, often fall short in capturing the breadth and depth of human multimodal communication, resulting in models that may not effectively generalize to varied real-world interactions. Moreover, existing frameworks like ZRIGF \cite{ref14}, which represent pioneering efforts in zero-resource image-grounded dialogue generation, do not seamlessly integrate into LLM architectures and depend on retrieving relevant images during inference to formulate responses. Thus, the challenge of enabling LLMs to generate multimodal dialogues in zero-resource scenarios, devoid of annotated multimodal dialogue datasets, remains unresolved.

To address the primary challenge of enabling LLMs to generate dialogues in zero-resource scenarios, we introduce the concept of implicit multimodal knowledge. Unlike existing approaches, which predominantly utilize explicit multimodal inputs such as images or sounds presented during interactions, we center on implicit multimodal knowledge. This form of knowledge, significantly distinct from the explicit forms commonly integrated in current models, refers to a mental imagery or conceptual understanding individuals have developed through their experiences \cite{ref15, ref16}. Implicit multimodal knowledge encompasses a broad spectrum of sensory, emotional, and contextual understandings that, although not directly observable, profoundly influence the nature of dialogues \cite{ref17, ref18}. By leveraging readily available large-scale image-text pair corpora to learn how to utilize implicit multimodal knowledge, it is possible to circumvent the issue of scarcity in high-quality, diverse multimodal dialogue datasets. However, current multimodal LLMs primarily engage with explicit multimodal inputs, focusing on direct responses to visual or auditory stimuli within dialogues \cite{ref19}, resulting in a lack of capability to incorporate such implicit knowledge. Therefore, the challenge becomes how to effectively distill and integrate implicit multimodal knowledge into LLMs, thereby significantly enhancing their ability to generate nuanced dialogues in zero-resource scenarios.

\begin{figure}
	\centering
	\includegraphics[scale=0.5]{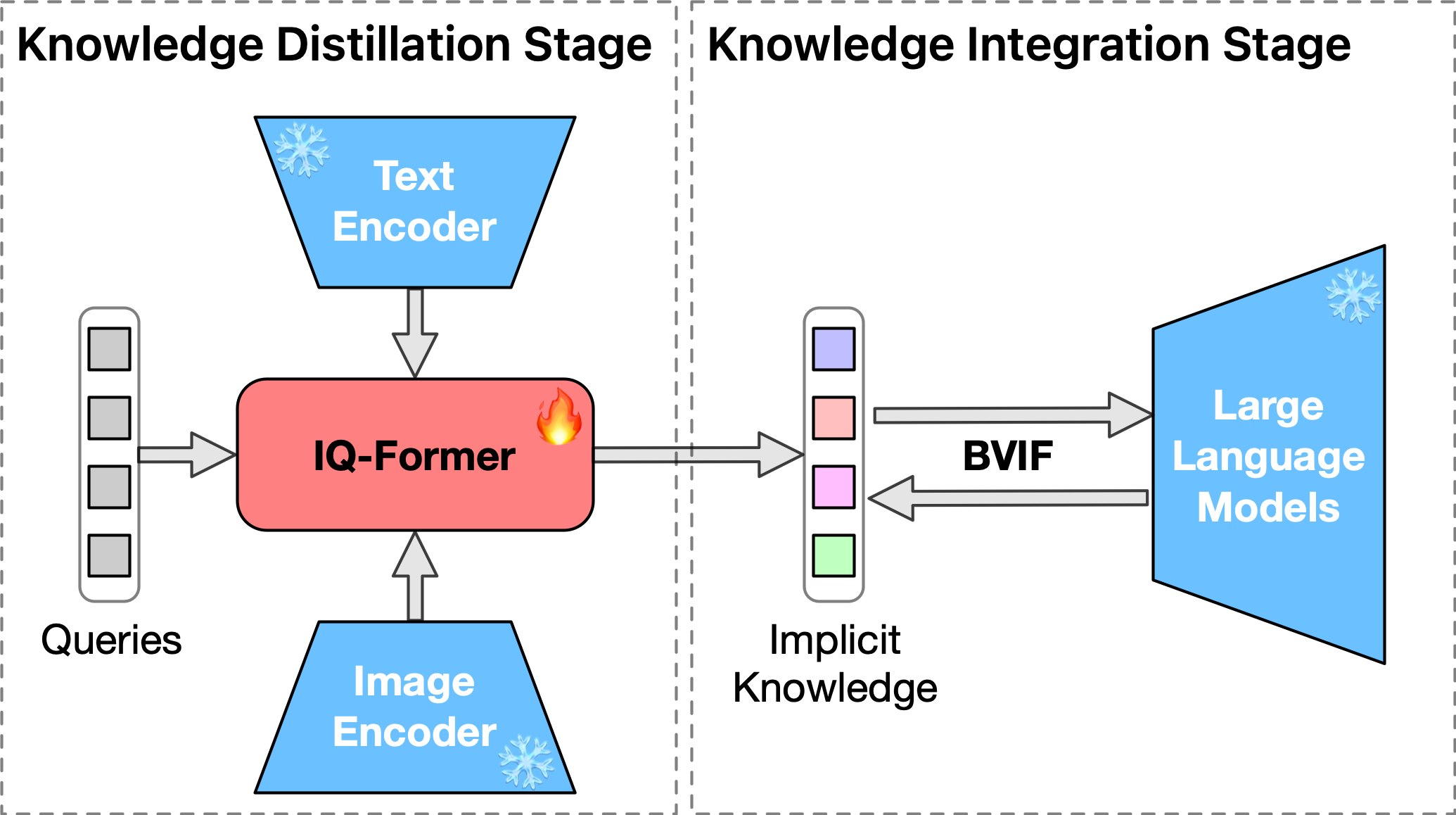}
	\caption{Intuition of our proposed approach.}
	\label{fig_overview}
\end{figure}

To navigate this refined challenge, we propose the Visual Implicit Knowledge Distillation Framework (VIKDF), a novel framework that reimagines the integration of multimodal knowledge into LLMs, focusing on the distillation and incorporation of visual implicit knowledge in a zero-resource scenario. VIKDF operates in two synergistic stages: knowledge distillation and knowledge integration, as illustrated in Figure \ref{fig_overview}. In the knowledge distillation stage, VIKDF utilizes a multimodal model that incorporates a text encoder, an image encoder, and a novel querying transformer termed the \textit{Implicit Query Transformer} (IQ-Former). This transformer, an advancement of the Q-Former \cite{ref7}, is specially tailored for extracting implicit knowledge. It employs a set of learnable query vectors to distill visual implicit knowledge from extensive image-text pair corpora. These vectors serve as the representation of the visual implicit knowledge, which can be then effectively integrated into LLMs. During the knowledge integration stage, we introduce a pioneering technique to seamlessly integrate the distilled visual implicit knowledge into LLMs, named \textit{Bidirectional Variational Information Fusion} (BVIF). BVIF leverages an instruction-aware dual-pathway approach to maximize the mutual information between textual context and distilled visual implicit knowledge, thereby capturing the essence of the visual implicit knowledge. This simultaneous optimization ensures coherent, context-rich dialogues and bridges the gap between explicit and implicit multimodal knowledge processing. Consequently, VIKDF enables LLMs to engage in complex dialogues without depending on annotated multimodal datasets, marking a significant step forward in zero-resource dialogue generation.

To validate our framework's efficacy, we conducted comprehensive experiments on the Image-Chat \cite{ref22} and Reddit Conversation datasets \cite{ref4}, benchmarking our method against several state-of-the-art baselines such as ChatGPT and ZRIGF. Through both automatic and human evaluations, VIKDF showcased its exceptional ability to fluently incorporate visual implicit knowledge into dialogues, thereby generating contextually rich, engaging, and coherent conversations, and outperforming existing models in zero-resource scenarios.

Our main contributions are highlighted as follows:
\begin{itemize}
	\item We propose a novel framework that distills and integrates visual implicit knowledge into LLMs, enabling them to generate more engaging dialogues without relying on any explicit images in zero-resource scenarios.
	\item We develop the Implicit Query Transformer and Bidirectional Variational Information Fusion techniques, effectively distilling and integrating visual implicit knowledge into LLMs and enhancing their dialogue generation capabilities.	
	\item We conduct extensive evaluations across two datasets in diverse scenarios, demonstrating the superior performance and robust generalization capabilities of VIKDF.
\end{itemize}

\section{Related Work}
\subsection{Multimodal Dialogue Generation}
Multimodal dialogue generation aims to produce responses that are natural and engaging, considering inputs from multiple modalities such as images, videos, or audio. This field requires models that have the ability to understand or generate content across these different modalities, leveraging this multimodal knowledge to enrich dialogues. Early research in this area \cite{ref20, ref21} was primarily focused on multimodal question-answering, where the goal was to respond to queries with inputs from various modalities. However, there has been a noticeable shift towards generating open-domain dialogues. In these cases, multimodal inputs serve to enrich conversations rather than strictly guide them, leading to two main streams of research.

The first stream focuses on dialogue generation based on multimodal information. In this approach, models utilize inputs such as images or videos to influence the dialogue generation process. These models typically employ multimodal encoders or attention mechanisms to integrate multimodal features with textual features, thus enhancing the relevance and diversity of the generated responses \cite{ref4, ref6, ref23}. This approach mimics face-to-face interactions where non-verbal cues influence but do not solely dictate the conversation. The second stream involves models that not only interpret multimodal inputs but also generate outputs across multiple modalities. This approach is akin to network-based dialogues, where communication often includes and sometimes relies on multimodal elements such as emojis, images, or videos. These models require more advanced capabilities for handling cross-modal generation and alignment, as well as ensuring the coherence and consistency of multimodal outputs \cite{ref24, ref25, ref26}.

Our research aligns with the first stream, aiming to enhance dialogue using multimodal inputs without generating multimodal outputs. However, most existing methodologies rely on annotated multimodal dialogue data, which is both scarce and expensive to obtain. In an effort to bridge this gap, Zhang \textit{et al.} \cite{ref14} introduced a zero-resource image-grounded framework that leverages images to enrich dialogues through a two-stage learning strategy. While innovative, this method requires access to relevant images during inference, which may not always be feasible. Our proposed approach differs by utilizing implicit multimodal knowledge, distilled from extensive collections of image-text pairs, to enhance dialogue generation. This strategy addresses the challenges of data scarcity and modality mismatch, enabling the generation of dialogues that are more natural, contextually rich, and authentically human-like.

\subsection{Multimodal Knowledge Distillation and Integration}
Multimodal knowledge distillation and integration are crucial for enabling LLMs to utilize multimodal information in dialogue generation. Distillation involves extracting and compressing information from a broad spectrum of multimodal data, such as image-text pairs. Integration, on the other hand, focuses on incorporating this distilled information into LLMs, thereby augmenting their capacity for understanding and generating multimodal dialogues.

Prior research on this topic has predominantly concentrated on explicit multimodal information, such as the direct fusion of image and textual features \cite{ref5, ref26, ref27, ref28} or employing pre-trained multimodal models \cite{ref29} to extract multimodal representations \cite{ref19, ref30}. However, these methods face limitations when applied to LLMs. Vision-Language Models (VLMs), like LLaVA \cite{ref53} and Mini-Gemini \cite{ref54}, are designed to process explicit multimodal inputs by interpreting both visual and textual modalities together. They typically train on large multimodal datasets to maximize alignment and shared understanding between vision and language components, enabling comprehensive tasks like visual question answering, image captioning, and scene understanding. Moreover, Li \textit{et al.} \cite{ref7} proposed an innovative solution, the Q-Former, which uses learnable query vectors to distill multimodal knowledge and integrate it into LLMs. Although this method has shown promising results in improving the multimodal capabilities of LLMs, it still encounters challenges in handling implicit multimodal knowledge, especially for dialogue generation.

Our work seeks to address a critical gap in multimodal dialogue generation by leveraging implicit multimodal knowledge to enhance LLMs' dialogue generation capabilities in zero-resource scenarios. This not only advances the technology of multimodal dialogue generation but also provides new insights into the complexities of human conversational interactions.

\section{Methodology}
\subsection{Task Formalization and Model Architecture}
The task of dialogue generation based on multimodal information is defined as generating a response $R$ based on a given dialogue context $C$ and corresponding multimodal information such as an image $I$. Our methodology diverges from conventional practices that leverage multimodal data directly, focusing instead on scenarios where such data is absent during both training and inference phases, known as zero-resource scenarios. Our strategy utilizes distilled visual implicit knowledge $K$, derived from $C$ via $P(K|C)$, to augment the quality of dialogue generation. Therefore, our objective is to generate $R$ by conditioning on both $C$ and $K$, formalized as $P(R|C, K)$. This leads to the model formulation:

\begin{equation}
  P(R|C) = P(R, K|C) = P(R|C, K)P(K|C)
\end{equation}
This formulation is justified as $K$ is inherently determined by $C$, explaining the rationale behind the first part of the equation.

To achieve this, we introduce a two-stage framework comprising knowledge distillation and integration, as shown in Figure \ref{fig_overview}. The framework is anchored by three principal components:

\subsubsection{Text Encoder and Image Encoder}
The text encoder and image encoder are responsible for encoding the textual and visual information from a large corpus of image-text pairs. We adopt the CLIP model \cite{ref29} as our text and image encoder, which is a pre-trained multimodal model that learns to align image and text features in a unified latent space. The text encoder transforms text input $T$ into a hidden representation $\mathbf{H}_T$, utilizing a transformer encoder architecture \cite{ref31}. Similarly, the image encoder, based on a vision transformer model \cite{ref32}, converts image input $I$ into a feature vector $\mathbf{H}_I$. The parameters of both encoders are set to remain unchanged.

\begin{figure}
	\centering
	\includegraphics[scale=0.5]{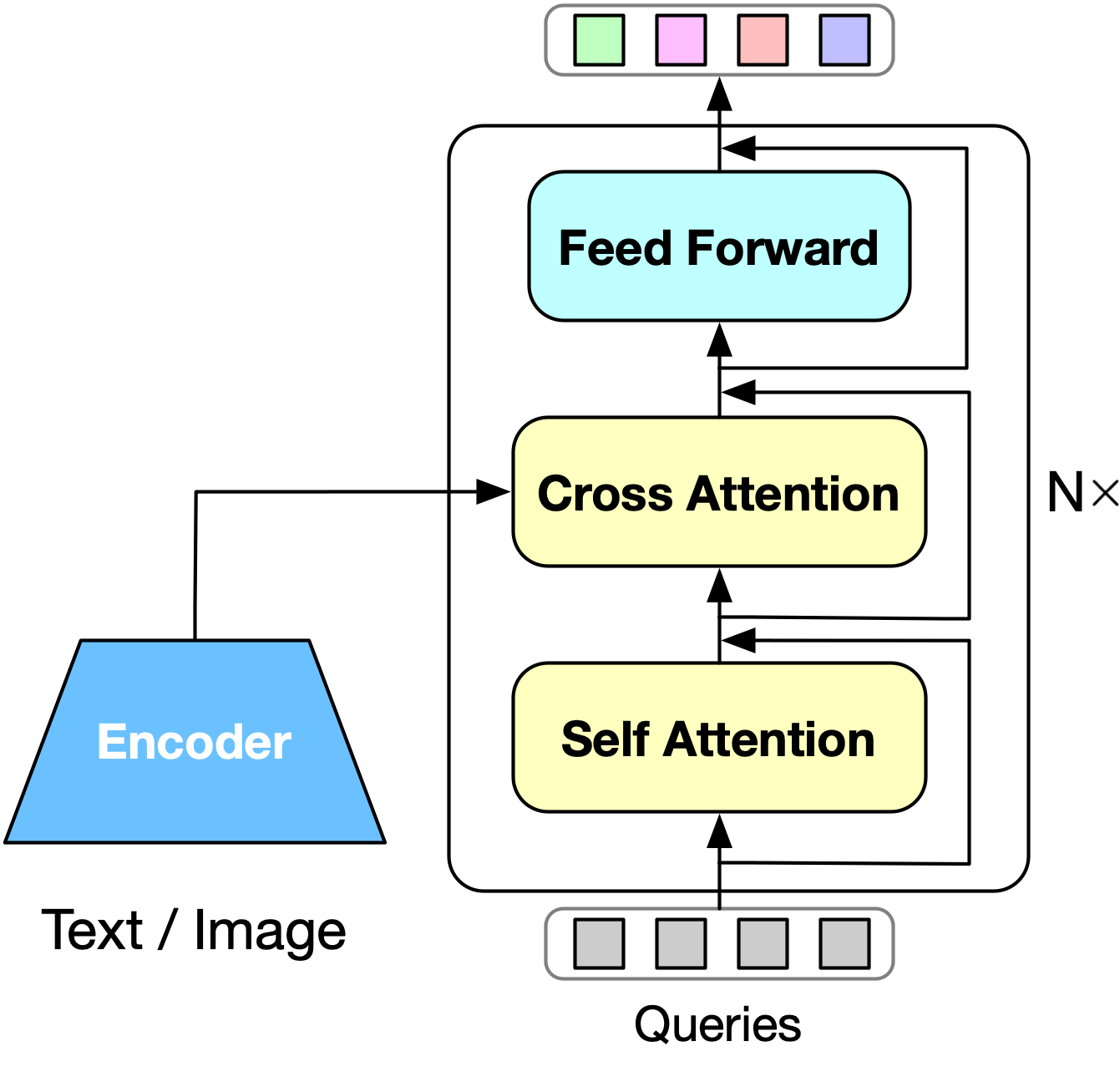}
	\caption{The architecture of IQ-Former.}
	\label{fig_if}
\end{figure}

\subsubsection{Implicit Query Transformer}
The IQ-Former is a specially designed querying transformer that distills visual implicit knowledge from encoded image-text pairs. As illustrated in Figure \ref{fig_if}, the IQ-Former is structured as a transformer encoder equipped with a cross-attention mechanism. Uniquely, its inputs are a set of learnable query vectors $\mathbf{Q} = {\mathbf{q}_1, \mathbf{q}_2, ..., \mathbf{q}_n}$, where $n$ is the number of queries. The IQ-Former uses the cross-attention mechanism to interact between the query vectors and both $\mathbf{H}_T$ and $\mathbf{H}_I$, and generates two sets of knowledge vectors $\mathbf{K}_T = {\mathbf{k}_{T_1}, \mathbf{k}_{T_2}, ..., \mathbf{k}_{T_n}}$ and $\mathbf{K}_I = {\mathbf{k}_{I_1}, \mathbf{k}_{I_2}, ..., \mathbf{k}_{I_n}}$, respectively. The vectors in $\mathbf{K}_T$ contain enriched textual information, whereas those in $\mathbf{K}_I$ are infused with explicit visual information. By optimizing the IQ-Former with specifically designed objectives, as described in Section \ref{section_kd}, it can be efficiently distill visual implicit knowledge into $\mathbf{K}_T$.

\subsubsection{Large Language Model}
The LLM is tasked with generating dialogue responses based on the dialogue context and the distilled visual implicit knowledge. We employ the Llama-2 model \cite{ref33} as our LLM, a pre-trained autoregressive language model renowned for its capability to produce natural and varied text. The LLM takes the dialogue context $C$ and distilled visual implicit knowledge $\mathbf{K}_T$ as inputs, and generates the response $R$ as output. To facilitate the integration of $\mathbf{K}_T$ into the LLM, we introduce a novel technique termed Bidirectional Variational Information Fusion, detailed further in Section \ref{section_ki}. The LLM's parameters are also frozen during training.

\begin{figure*}
	\centering
	\includegraphics[scale=0.5]{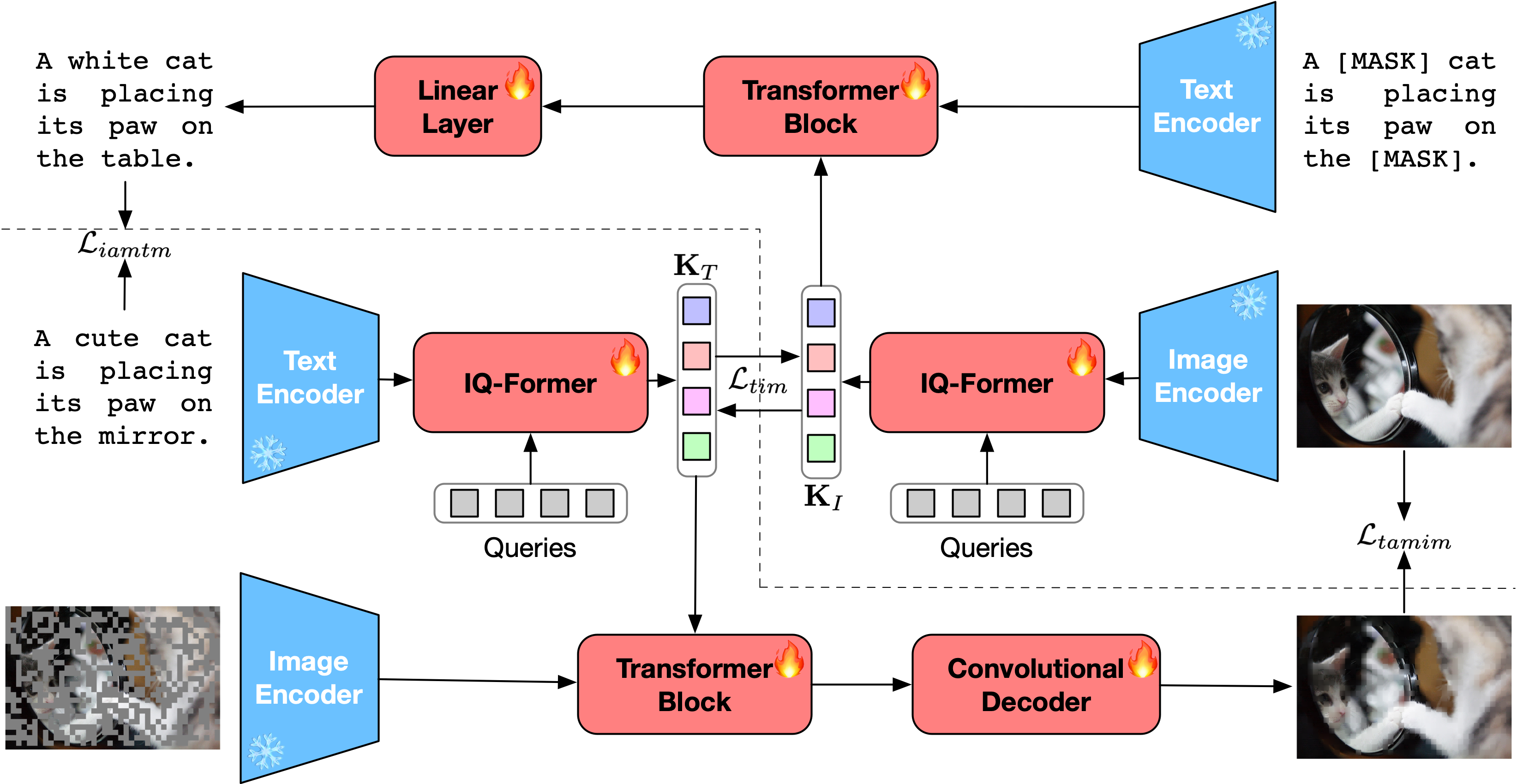}
	\caption{Overview of the knowledge distillation stage. The central part illustrates Text-Image Matching. Below the dashed line lies Text-Assisted Masked Image Modeling, and above, Image-Assisted Masked Text Modeling.}
	\label{fig_kd}
\end{figure*}

\subsection{Knowledge Distillation Stage}
\label{section_kd}
The goal of the knowledge distillation stage is to develop a model, denoted as $P(K|C)$, capable of deducing visual implicit knowledge $K$ from a given dialogue context $C$. To achieve this, we use a large corpus of image-text pairs to optimize IQ-Former so that the knowledge vectors $\mathbf{K}_T$ can encapsulate visual implicit knowledge pertinent to the input context. As depicted in Figure \ref{fig_kd}, the IQ-Former distills the visual implicit knowledge into $\mathbf{K}_T$ by using three objectives.

\subsubsection{Text-Image Matching}
The text-image matching objective ensures the alignment of knowledge vectors $\mathbf{K}_T$ and $\mathbf{K}_I$ within a shared latent space, promoting consistency and coherence between the knowledge extracted from both text and image. We employ a contrastive learning strategy \cite{ref29} to achieve this, by maximizing the cosine similarity between $\mathbf{K}_T$ and $\mathbf{K}_I$ for matching pairs and minimizing the cosine similarity for non-matching pairs. The loss for text-image matching is formulated as follows:
\begin{equation}
\begin{aligned}
\mathcal{L}_{tim} = -\sum_{i=1}^{N} (& \log \frac{\exp(\cos({\mathbf{K}_T}_i, {\mathbf{K}_I}_i) / \tau)}{\sum_{j=1}^{N} \exp(\cos({\mathbf{K}_T}_i, {\mathbf{K}_I}_j) / \tau)} \\
 + \ & \log \frac{\exp(\cos({\mathbf{K}_T}_i, {\mathbf{K}_I}_i) / \tau)}{\sum_{j=1}^{N} \exp(\cos({\mathbf{K}_T}_j, {\mathbf{K}_I}_i) / \tau)})
\end{aligned}
\end{equation}
where $\tau$ is a learnable temperature parameter that controls the distribution's concentration. By minimizing this loss, the IQ-Former learns to generate knowledge vectors that are semantically similar to the corresponding image features, thereby encapsulating visual implicit knowledge from the text input.

\subsubsection{Text-Assisted Masked Image Modeling}
The objective of text-assisted masked image modeling is to reconstruct the masked portions of an image input, denoted as $\mathbf{I}$, utilizing text knowledge vectors $\mathbf{K}_T$. This process enables the IQ-Former to apply textual information to deduce missing visual data, thereby improving the extraction of visual implicit knowledge. Initially, a certain percentage of pixels in $\mathbf{I}$ is randomly masked, producing a masked image $\mathbf{I}'$. This image is then fed into the image encoder to obtain the masked image feature vectors $\mathbf{H}_{I'}$. Subsequently, $\mathbf{H}_{I'}$ is combined with $\mathbf{K}_T$ using a transformer block that includes multi-head attention and a feed-forward network. The resulting output vectors $\mathbf{O}_{I'}$, now enriched with textual information, are input into a convolutional decoder consisting of a convolution layer followed by a pixel-shuffle operation to generate the reconstructed image $\hat{\mathbf{I}}$. The loss for text-assisted masked image modeling, expressed as the mean absolute error between $\mathbf{I}$ and $\hat{\mathbf{I}}$ in the masked regions, is defined as follows:
\begin{equation}
  \mathcal{L}_{tamim} = \frac{1}{N_i} \sum_{i=1}^{N_i} |\mathbf{I}_i - \hat{\mathbf{I}}_i|
\end{equation}
where $N_i$ represents the count of masked pixels, with $\mathbf{I}_i$ and $\hat{\mathbf{I}}_i$ being the pixel values of the original and reconstructed images, respectively. Minimizing this loss enables the IQ-Former to create knowledge vectors that contain sufficient visual implicit knowledge to assist the image reconstruction.

\subsubsection{Image-Assisted Masked Text Modeling}
The goal of image-assisted masked text modeling is to recover the masked tokens in a text input $T$ using image knowledge vectors $\mathbf{K}_I$. This objective encourages the IQ-Former to use visual information to deduce missing textual content, enhancing cross-modal knowledge alignment. This objective is similar to the masked language modeling task in BERT \cite{ref34}, but with the addition of visual information. Similar to the text-assisted masked image modeling, we can obtain the output vectors $\mathbf{O}_{T'}$ that incorporates visual information. To reconstruct masked tokens, $\mathbf{O}_{T'}$ is fed into a linear layer followed by a softmax function to predict the vocabulary's probability distribution for each masked token. 
The loss for image-assisted masked text modeling, calculated as the cross-entropy loss between the predicted distribution and the true masked tokens, is:
\begin{equation}
  \mathcal{L}_{iamtm} = -\frac{1}{N_t} \sum_{i=1}^{N_t} \log P(T_i|\mathbf{O}_T')
\end{equation}
where $N_t$ is the total number of masked tokens, and $P(T_i|\mathbf{O}_T')$ is the predicted probability of the original token $T_i$ based on the output vectors $\mathbf{O}_T'$. By minimizing this loss, the IQ-Former is trained to produce output vectors that contain sufficient cross-modal knowledge to assist the text reconstruction, thus capturing cross-modal knowledge alignment.

The comprehensive loss function for the knowledge distillation stage is the sum of the losses from the three objectives, weighted by their respective importance:
\begin{equation}
  \mathcal{L}_{kd} = \lambda_1 \mathcal{L}_{tim} + \lambda_2 \mathcal{L}_{tamim} + \lambda_3 \mathcal{L}_{iamtm}
  \label{eq_kd}
\end{equation}
where $\lambda_1$, $\lambda_2$, and $\lambda_3$ are hyperparameters that determine the significance of each objective. By minimizing this loss, the IQ-Former is trained to effectively distill and encapsulate visual implicit knowledge derived from image-text pairings into the knowledge vectors $\mathbf{K}_T$. These vectors can then be used as the input $K$ for the LLM during the knowledge integration stage.

\begin{figure*}
	\centering
	\includegraphics[scale=0.5]{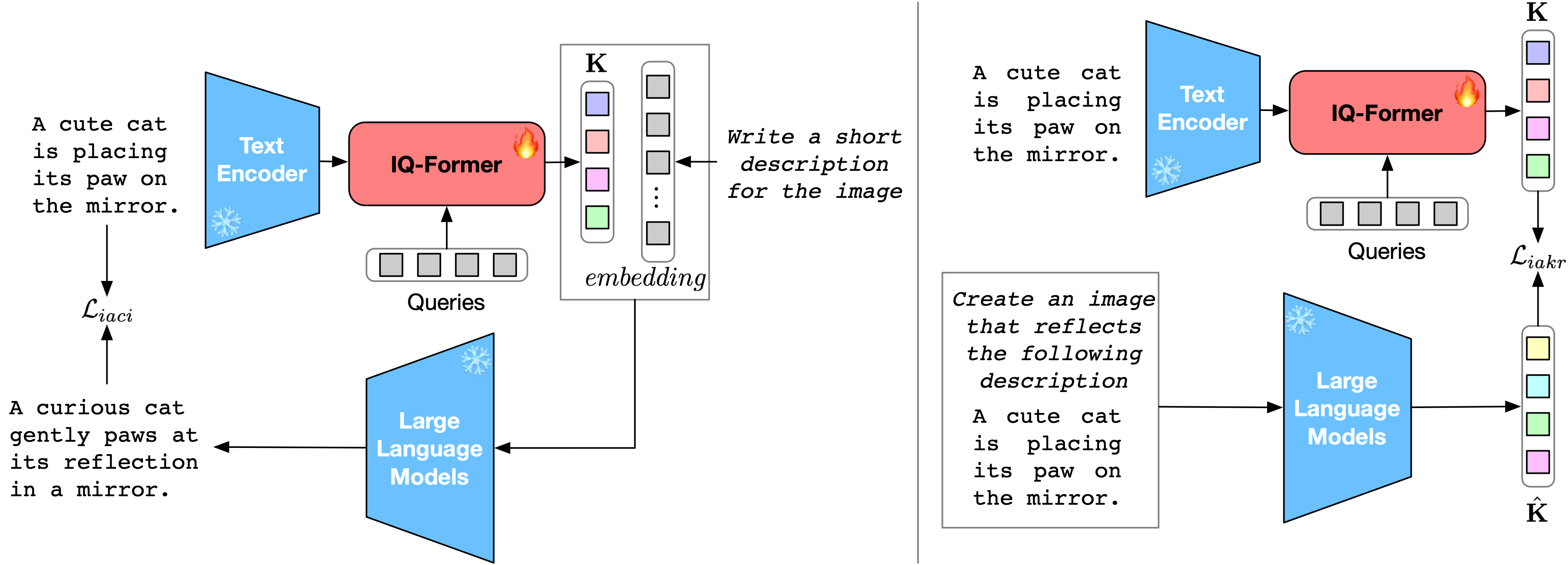}
	\caption{Overview of the knowledge integration stage. Instruction-aware Contextual Inference on the left and Instruction-aware Knowledge Reconstruction on the right}
	\label{fig_ki}
\end{figure*}

\subsection{Knowledge Integration Stage}
\label{section_ki}
The objective of the knowledge integration stage is to train a model, denoted as $P(R|C, K)$, that generates a dialogue response $R$ informed by the dialogue context $C$ and distilled visual implicit knowledge $K$. This process leverages image-text pair data to optimize the learnable knowledge vectors $\mathbf{K}$, referred to earlier as $\mathbf{K}_T$, thus equipping the LLM with the ability to comprehend and incorporate the knowledge encapsulated in $\mathbf{K}$. As depicted in Figure \ref{fig_ki}, the LLM integrates visual implicit knowledge through a pioneering technique named Bidirectional Variational Information Fusion. BVIF utilizes an instruction-aware dual-pathway approach, with each path providing a distinct yet complementary mechanism for knowledge fusion.

\subsubsection{Instruction-aware Contextual Inference}
The instruction-aware contextual inference pathway aims to enable the LLM to decode and integrate the textual intricacies contained within the distilled visual implicit knowledge. This pathway introduces distilled visual implicit knowledge $K$ as soft visual prompts, directing the LLM towards generating text $T$ that aligns with the visual context. Initially, we attach the text encoder to the IQ-Former and freeze its parameters. Subsequently, the IQ-Former is connected to the LLM, using a linear layer to project the knowledge vectors $\mathbf{K}$ into the same dimension as the LLM's text embedding. These projected query embeddings are then positioned at the beginning of the input text embeddings sequence for the LLM. Additionally, we use a set of text-based prompts, such as ``\textit{Write a short description for the image.}'', to fine-tune the LLM's generation according to the specific task. The LLM then generates text $T$ by optimizing the likelihood of predicting each subsequent token, based on preceding tokens, the query embeddings, and the text prompts. The instruction-aware contextual inference loss is formally defined as:
\begin{equation}
  \mathcal{L}_{iaci} = -\frac{1}{N_t} \sum_{i=1}^{N_t} \log P(T_i|K, P, T_{<i})
\end{equation}
where $N_t$ represents the text's token count, $P$ the text prompt, and $P(T_i|K, P, T_{<i})$ the probability of token $T_i$ conditioned on the knowledge $K$, prompt $P$, and preceding tokens $T_{<i}$. Minimizing this loss instructs the LLM to produce text reflective of the visual implicit knowledge distilled by the IQ-Former from the image-text pairs.

\subsubsection{Instruction-aware Knowledge Reconstruction}
The instruction-aware knowledge reconstruction pathway is a crucial component of the BVIF technique, aimed at augmenting the LLM's proficiency in interpreting and utilizing distilled visual implicit knowledge $K$ for dialogue generation. This pathway focuses on reconstructing knowledge $K$ from the generated text $T$, thereby establishing a bidirectional information flow between the text and the visual implicit knowledge.

A primary challenge in this task is to ensure that the knowledge $K$ is deeply embedded and reflected in the generated text $T$. While the instruction-aware contextual inference pathway can mitigate this challenge to an extent, fully embedding visual implicit knowledge into the LLM's learning process remains challenging. To address this, we implement a mutual information maximization mechanism to quantify the dependency between $K$ and $T$. The mutual information, denoted as $I(K, T)$, is defined as the expected value of the logarithmic ratio between the joint probability distribution of $K$ and $T$ and the product of their marginal probability distributions. However, direct optimization of $I(K, T)$ is intractable due to computational complexity. Consequently, we maximize a lower bound of $I(K, T)$ through a variational information maximization approach, as detailed in \cite{ref35}. This approach is mathematically represented as:
\begin{equation}
  I(K, T) \geq \mathbb{E}_{p(K)} \mathbb{E}_{p(T|K)} \log q_{\phi}(K | T)
\end{equation}
where $q_{\phi}(K | T)$ is a variational approximation of the posterior probability of $K$ given $T$, parameterized by $\phi$. This formulation allows for the approximation of mutual information by learning a function $q_{\phi}$ that predicts $K$ from $T$.

In practice, we use the LLM as the function $q_{\phi}$ that infers $K$ based on $T$. Specifically, we feed the generated text $T$ and a text prompt, such as ``\textit{Create an image that reflects the following description:}'', into the LLM to yield output vectors $\mathbf{O}_K$. A linear layer then projects $\mathbf{O}_K$ to match the dimensionality of the original knowledge vectors $\mathbf{K}$. The instruction-aware knowledge reconstruction loss is calculated as the mean squared error between the original knowledge vectors $\mathbf{K}$ and the reconstructed knowledge vectors $\hat{\mathbf{K}}$ across all queries:
\begin{equation}
  \mathcal{L}_{iakr} = \frac{1}{N_q} \sum_{i=1}^{N_q} (\mathbf{K}_i - \hat{\mathbf{K}}_i)^2
\end{equation}
where $N_q$ is the total number of queries, with $\mathbf{K}_i$ and $\hat{\mathbf{K}}_i$ representing the original and reconstructed knowledge vectors, respectively. Minimizing this loss enables the LLM to produce knowledge vectors consistent with the generated text, thus reinforcing a bidirectional flow of information between the text and the visual implicit knowledge.

The overall loss function for the knowledge integration stage is the weighted sum of the two objectives:
\begin{equation}
  \mathcal{L}_{ki} = \lambda_4 \mathcal{L}_{iaci} + \lambda_5 \mathcal{L}_{iakr}
  \label{eq_ki}
\end{equation}
where $\lambda_4$ and $\lambda_5$ are hyperparameters that control the relative importance of each objective. By minimizing this loss, the LLM is trained to generate dialogue responses that are coherent and engaging based on both the dialogue context and the distilled visual implicit knowledge.

\subsection{Zero-Resource Learning Detail}
\subsubsection{Training}
To train our framework, we initially use four large-scale image-text datasets, including COCO Captions \cite{ref36}, CC3M \cite{ref37}, CC12M \cite{ref38}, and SBU \cite{ref39}, for pre-training the IQ-Former and aligning it with the LLM. We follow the same data processing and augmentation methods as BLIP-2 \cite{ref7}, which generates synthetic captions for web images using a pre-trained captioning model and a CLIP model. These datasets contain tens of millions of image-text pairs that cover a wide range of topics and scenarios, providing a rich source of visual implicit knowledge. We concurrently train both stages of our framework, employing image-text pair data in accordance with Equations \eqref{eq_kd} and \eqref{eq_ki}.

For fine-tuning on specific dialogue tasks, we utilize only textual data without any image input, aligning with our goal of zero-resource scenarios where multimodal data is unavailable during fine-tuning and inference. During this phase, the IQ-Former distills visual implicit knowledge $K$ from the dialogue context $C$ alone by modeling $P(K|C)$. The fine-tuning process optimizes the negative log-likelihood loss to enhance the probability $P(R|C, K)$, as detailed in Section \ref{section_infer}.

\subsubsection{Inference}
\label{section_infer}
To perform zero-resource inference, we leverage the trained IQ-Former and the LLM to produce dialogue responses enriched with the visual implicit knowledge distilled from the dialogue context. Given a dialogue context $C$, we first pass it through the IQ-Former with the text encoder to model $P(K|C)$, thereby acquiring knowledge vectors $K$ that encapsulate visual implicit knowledge derived from $C$. Notably, no image input is provided at this stage; the IQ-Former operates solely on the textual input $C$. Next, we input $K$ and $C$ into the LLM, which models $P(R|C,K)$, and then generate the response $R$ by sampling from the probability distribution over the vocabulary. Since the LLM has learned to integrate the visual implicit knowledge into the dialogue generation during training, it can produce natural and engaging responses that are consistent with the visual context implied by $C$, even in the absence of any explicit multimodal inputs.

\section{Experiments}
In this section, we evaluate the performance of our proposed framework, VIKDF, on the task of zero-resource dialogue generation. We benchmark VIKDF against a range of baselines and ablation models, conducting both automatic and human evaluations. Additionally, we present qualitative examples to demonstrate the effectiveness of our approach.

\subsection{Datasets}
We employ two datasets to evaluate our framework: Image-Chat \cite{ref22} and Reddit Conversation \cite{ref4}. The Image-Chat dataset, crucial for validating VIKDF's efficacy in zero-resource scenarios, is a large-scale image-grounded dialogue dataset featuring 202,000 dialogues across 202,000 images. Each dialogue comprises a single turn of context and response, with the latter influenced by the corresponding image. This dataset is divided into 186,782 training dialogues, 5,000 validation dialogues, and 9,997 testing dialogues. The Reddit Conversation dataset is served to assess the performance of visual implicit knowledge. This dataset, sourced from the widely-used online forum, encompasses an extensive variety of dialogue topics and styles. It has been preprocessed to include 1,000,000 training dialogues, with an additional 20,000 for validation and 20,000 for testing.

Similar to \cite{ref14}, we fine-tune the model using the Reddit Conversation dataset to enhance the foundational dialogue generation capabilities of the LLM. To evaluate the model's zero-resource dialogue generation capabilities, we directly perform inference on the Image-Chat dataset after fine-tuning on Reddit Conversation. Additionally, we conduct experiments by fine-tuning the model on Image-Chat without providing any images.

\subsection{Implementation Details}
The VIKDF implementation utilizes the Hugging Face Transformers library \cite{ref22}. For the text and image encoders, we initialize them with the pre-trained CLIP model that employs a ViT-L/14 Transformer architecture. The IQ-Former is initialized with the pre-trained BERT-base model, while the chat version of Llama-2 with 7B parameters is used as the large language model. To ensure seamless integration and information flow across models of varying dimensionalities, the necessary linear transformations are applied, which are not mentioned in the methodology.

VIKDF adopts a simultaneous training regimen for both knowledge distillation and integration stages. This strategy is executed on four NVIDIA RTX 4090 GPUs, utilizing a batch size of 128 across 100,000 training steps. For optimization, we employ mixed-precision training with bfloat16 and utilize the AdamW optimizer \cite{ref40} with a learning rate of $1e^{-4}$. The learning rate is accompanied by a weight decay of 0.05 and a linear learning rate warmup over the first $10\%$ of the total steps. The hyperparameters $\lambda_1$ through $\lambda_5$ are set as 0.5, 0.2, 0.2, 1, and 0.5, respectively. For the set of query vectors within the IQ-Former, we set $n$ to 32, with each vector having a dimensionality of 768. In the text-assisted masked image modeling, we apply a random masking strategy with $28 \times 28$ patches at a 0.6 ratio. For the image-assisted masked text modeling, the strategy involves a 0.15 mask ratio.

\begin{table*}
  \caption{Assessment of automated metrics: $\dag$ denotes a zero-resource scenario with no annotated images, while $\ddag$ indicates a fully zero-resource scenario without any prior training on task-specific datasets. Bold font highlights the best performance in each column, and underlines signify the second-best performance. Significant improvements are marked by $^\star$ ($p < 0.05$).}
  \label{tab_automatic}
  \centering
  \resizebox{.86\textwidth}{!}{
  \begin{tabular}{clr*{9}{c}}
    \toprule
	\multirow{2}{*}{Task} & \multirow{2}{*}{Methods} & \multirow{2}{*}{\makecell{\#Tuned \\ Params}} & \multirow{2}{*}{PPL} & \multirow{2}{*}{BLEU-1} & \multirow{2}{*}{ROUGE-L} & \multirow{2}{*}{Average} & \multirow{2}{*}{Extrema} & \multirow{2}{*}{Greedy} & \multirow{2}{*}{Dis-1} & \multirow{2}{*}{Dis-2} \\
	& & & & & & & & & & \\
    \midrule
    \multirow{8}{*}{\makecell{Reddit\\ Conversation}}
    & Seq2Seq & 24M & 77.27 & 12.21 & 10.81 & 78.38 & 40.06 & 62.64 & 0.53 & 1.96 \\
    & BART & 406M & 44.73 & 13.51 & 12.50 & 80.21 & 41.63 & 63.72 & 4.17 & 16.98 \\
    & Llama-2$^\ddag$ & 0M & 155.69 & 10.27 & 10.52 & 81.72 & 35.95 & 60.70 & 4.94 & 31.19 \\
    & ChatGPT$^\ddag$ & 0M & {-} & 11.62 & 11.29 & \underline{82.39} & 37.48 & 62.05 & 5.28 & \textbf{38.63}$^\star$ \\
    \cmidrule(lr){2-11}
    & ImgVAE & - & 72.06 & 12.58 & 12.05 & 79.95 & 42.38 & 63.55 & 1.52 & 6.34 \\
    & Maria & 135M & 56.23 & 14.10 & 12.66 & 81.76 & 43.04 & 63.98 & 4.83 & 22.87 \\
    & ZRIGF & 658M & \underline{36.21} & \underline{16.06} & \underline{14.51} & 82.27 & \underline{43.79} & \underline{64.53} & \underline{5.79} & 26.57 \\
    \cmidrule(lr){2-11}
    & VIKDF & 159M & \textbf{14.98}$^\star$ & \textbf{16.47}$^\star$ & \textbf{14.84}$^\star$ & \textbf{82.54}$^\star$ & \textbf{43.84} & \textbf{64.88}$^\star$ & \textbf{6.53}$^\star$ & \underline{35.12} \\
    \midrule
    \multirow{14}{*}{Image-Chat}
    & Seq2Seq & 24M & 50.82 & 11.34 & 13.65 & 82.95 & 47.45 & 65.67 & 1.28 & 7.80 \\
    & BART & 406M & 37.26 & 13.41 & 14.24 & 84.48 & 48.57 & 66.49 & 2.44 & 15.79 \\
    & Llama-2$^\ddag$ & 0M & 193.20 & 9.93 & 11.56 & 85.44 & 40.18 & 63.05 & 4.69 & 30.81 \\
    & ChatGPT$^\ddag$ & 0M & {-} & 10.77 & 11.62 & 86.17 & 43.02 & 64.66 & 5.32 & \textbf{37.77}$^\star$ \\
    \cmidrule(lr){2-11}
    & ImgVAE & - & 41.94 & 16.07 & 15.98 & 85.81 & 49.59 & 67.44 & 1.68 & 7.22 \\
    & Maria$^\dag$ & 135M & 37.49 & 14.74 & 14.59 & 85.72 & 50.58 & 66.89 & 2.57 & 11.99 \\
    & Maria$_{zero}^\ddag$ & 0M & 135.49 & 11.75 & 12.13 & 83.51 & 45.57 & 64.48 & 1.89 & 7.32 \\
	& ZRIGF$^\dag$ & 658M & 29.82 & 16.86 & \underline{17.21} & \underline{86.30} & \textbf{51.41}$^\star$ & \textbf{68.56} & 2.59 & 10.62 \\
    & ZRIGF$_{1/4}^\dag$ & 658M & 35.41 & 16.35 & 16.59 & 85.75 & 49.95 & 67.20 & 4.61 & 22.66 \\
    & ZRIGF$_{zero}^\ddag$ & 0M & 105.12 & 15.17 & 15.13 & 84.52 & 45.95 & 65.70 & 5.25 & 29.38 \\
    \cmidrule(lr){2-11}
    & VIKDF$^\dag$ & 159M & \textbf{12.54}$^\star$ & \textbf{17.93}$^\star$ & \textbf{17.45}$^\star$ & \textbf{86.67}$^\star$ & \underline{50.88} & \underline{68.53} & 4.41 & 23.05 \\    
    & 1/4 Data$^\dag$ & 159M & \underline{12.84} & \underline{17.81} & 16.91 & 86.07 & 50.01 & 67.55 & 4.40 & 21.53 \\
    & 1/8 Data$^\dag$ & 159M & 13.12 & 17.70 & 16.84 & 85.98 & 49.76 & 67.35 & 4.28 & 20.54 \\
    & Zero Data$^\ddag$ & 0M & 27.32 & 15.35 & 15.30 & 85.04 & 45.86 & 66.02 & \textbf{6.17}$^\star$ & \underline{32.93} \\ 
    \bottomrule
  \end{tabular}
  }
\end{table*}

\subsection{Baseline Models}
We compare VIKDF with the following baseline models:
\begin{itemize}
	\setlength{\itemsep}{0pt}
	\setlength{\parskip}{0pt}
	\item Seq2Seq \cite{ref41}, a foundational architecture for sequence-to-sequence processing that includes an encoder and a decoder equipped with Long Short-Term Memory (LSTM) units. 
	\item BART \cite{ref42}, a pre-eminent sequence-to-sequence pre-training model that utilizes Transformer architecture.
	\item ImgVAE \cite{ref4}, which employs variational autoencoder technology to integrate visual information into dialogue generation.
	\item Maria \cite{ref5}, a visual experience-powered conversational agent that enriches dialogues with experiences from the visual world through a large-scale image index.
	\item Llama-2 \cite{ref33}, the LLM serving as our framework's backbone. It is a pre-trained autoregressive model capable of generating natural and diverse text.
	\item ChatGPT \cite{ref2} by OpenAI, which leverages the GPT architecture to generate human-like text responses, incorporating a mix of supervised and reinforcement learning techniques for dialogue systems. We use the gpt-3.5-turbo version in this paper.
	\item ZRIGF \cite{ref14}, a state-of-the-art model for zero-resource image-grounded dialogue generation that combines multimodal learning with a two-stage strategy.
\end{itemize}

Among them, ImgVAE, Maria, and ZRIGF are multimodal models, whereas the rest are unimodal text-based models. We have implemented all methods except for ImgVAE, where we directly used the outcomes from \cite{ref4}. We prompt Llama-2 following the approach outlined in the Hugging Face blog\footnote{\url{https://huggingface.co/blog/llama2}}, and employ ChatGPT in accordance with the methodology described in \cite{ref14}.

\subsection{Evaluation Metrics}
In accordance with \cite{ref5} and \cite{ref14}, we use both automatic and human evaluation metrics to assess the performance of VIKDF and the baseline models. 

For automatic evaluation, we employ the following metrics: (1) \textbf{Perplexity} (PPL) measures the model's fluency, with lower values indicating better performance. (2) \textbf{BLEU-1} \cite{ref43} and \textbf{ROUGE-L} \cite{ref44} evaluate the alignment of generated responses with human references, focusing on word-level accuracy and sequence similarity, respectively. (3) For semantic analysis, we employ \textbf{Average}, \textbf{Extrema} and \textbf{Greedy} metrics \cite{ref45} to measure the cosine similarity between word embeddings of generated and reference texts, capturing semantic coherence. (4) \textbf{Dis-1} and \textbf{Dis-2} metrics \cite{ref46} quantify the diversity of the model's output by calculating the uniqueness of unigrams and bigrams, respectively, ensuring the model's capability to produce varied and engaging responses. 

For human evaluation, we engage three evaluators to collect ratings from human annotators. We randomly sample 100 dialogues from the test set, and ask three evaluators to rate each dialogue on a scale of 1 to 5, based on the following criteria: (1) \textbf{Relevance}: How relevant and related is the generated response to the given context and image? (2) \textbf{Informativeness}: How much new and useful information does the generated response provide in the context of the dialogue? (3) \textbf{Fluency}: How natural, readable, and grammatically correct is the generated response? The final score for each criterion is computed as the average rating across all evaluators. To ensure the reliability of the evaluation process and measure the agreement among evaluators, Fleiss' \textbf{Kappa} \cite{ref47} statistic is applied to evaluate the concordance among evaluators.

\section{Results and Discussion}
Our evaluations differentiate between models operating in distinct zero-resource scenarios, as denoted by $\dag$ and $\ddag$. The $\dag$ symbol signifies scenarios where models operate without using annotated images during training and inference, while $\ddag$ denotes a fully zero-resource condition, in which models generate dialogues without any prior training on task-specific datasets.

\subsection{Automatic Evaluation Results}
Our proposed VIKDF demonstrates outstanding performance in zero-resource dialogue generation, outperforming both traditional and multimodal baseline models across various metrics on the Reddit Conversation and Image-Chat datasets. To ensure the robustness and reliability of our findings, we conducted additional experiments and statistical significance tests. Specifically, we ran VIKDF five times with different random seeds and performed one-sample t-tests comparing VIKDF to the best-performing baseline for each metric. Table \ref{tab_automatic} presents a comprehensive comparison of the automated evaluation metrics.

In the Reddit Conversation task, VIKDF achieves a significantly lower perplexity of 15.01, indicating superior fluency in generated dialogues compared to the nearest competitor, ZRIGF, which records a PPL of 36.21 in scenarios with explicit image guidance. Moreover, VIKDF outperforms all baselines in BLEU-1 and ROUGE-L scores, achieving 16.41 and 14.82, respectively, which highlights its ability to generate responses that are closely aligned with human references. Additionally, VIKDF surpasses other models in both Average and Greedy semantic similarity metrics, underscoring its enhanced ability to maintain semantic coherence in dialogues. Despite scoring slightly lower on the Extrema metric compared to ZRIGF, VIKDF remains competitive. Although it has a marginally lower Dis-2 score than ChatGPT, VIKDF's strong performance in generating diverse dialogues is evident. This suggests that while maintaining high relevance and accuracy, VIKDF also generates a wide range of responses, contributing to more dynamic and engaging dialogues. Notably, VIKDF, with only 159M tuned parameters, significantly outperforms ZRIGF, which has 658M tuned parameters, in 7 out of 8 evaluation metrics. This indicates that VIKDF utilizes its parameters more efficiently, resulting in more coherent and contextually appropriate responses. Furthermore, VIKDF achieves superior performance compared to models that depend on explicit images, highlighting its effectiveness in enhancing dialogue capabilities without direct visual inputs. The outstanding performance of VIKDF in the Reddit Conversation task highlights the substantial benefits of incorporating implicit knowledge into purely text-based dialogue systems, proving that a deep fusion of visual and contextual understanding significantly enhances the quality and engagement of the dialogues generated.

In the Image-Chat task, VIKDF's capabilities are examined under various conditions, including limited data availability scenarios (1/4 and 1/8 of the full training set) and fully zero-resource scenarios. We can see that VIKDF showcases superior performance in various zero-resource scenarios, especially notable in the absence of annotated images ($\dag$). Even without explicit images, VIKDF's performance notably surpasses that of the state-of-the-art model ZRIGF in almost all metrics, except for Extrema and Greedy, where it still achieves second-best performance. The t-test results confirm that these improvements are not due to random chance. Additionally, VIKDF demonstrates resilience by maintaining stable performance despite substantial reductions in training data, highlighting its robustness and generalization ability. This is in stark contrast to other models, which exhibit significant performance drops when transitioning from full data to zero data scenarios. Remarkably, VIKDF maintains its competitive edge even in fully zero-resource conditions ($\ddag$), showcasing its exceptional capability to generate diverse, relevant, and engaging dialogues without task-specific training data. The achievements of VIKDF on the Image-Chat dataset affirm its leading position in zero-resource dialogue generation, illustrating unmatched adaptability and advanced integration of visual implicit knowledge.

\begin{table}
  \caption{Human evaluation outcomes for the Image-Chat dataset in a fully zero-resource scenario.}
  \label{tab_human}
  \centering
  \resizebox{.45\textwidth}{!}{
  \begin{tabular}{l *{4}{c}}
    \toprule
    {Methods} & {Relevance} & {Informativeness} & {Fluency} & {Kappa} \\
    \midrule
    Llama-2 & 2.94 & 2.75 & 4.27 & 0.55 \\
    ChatGPT & 3.22 & \underline{3.28} & \textbf{4.50} & 0.60 \\
    ZRIGF & \underline{3.55} & 3.15 & 4.20 & 0.57 \\
    VIKDF & \textbf{3.84} & \textbf{3.30} & \underline{4.34} & 0.59 \\
    \bottomrule
  \end{tabular}
  }
\end{table}

\subsection{Human Evaluation Results}
Table \ref{tab_human} presents the human evaluation outcomes for the Image-Chat dataset within a fully zero-resource scenario ($\ddag$), comparing the performance of VIKDF against Llama-2, ChatGPT, and ZRIGF. Given that the human evaluation was conducted using a 5-point Likert scale, even a small absolute improvement in scores should be considered significant. Notably, VIKDF achieves the highest scores in both Relevance and Informativeness, indicating its exceptional ability to generate dialogues that are not only closely related to the given context and images but also rich in informative content. Specifically, VIKDF achieves an 8.17\% improvement in Relevance, a 4.76\% improvement in Informativeness, and a 3.33\% improvement in Fluency over ZRIGF. Although ChatGPT receives the highest rating in Fluency, VIKDF remains competitive, with a score of 4.34, illustrating its capacity to produce responses that are natural, coherent, and grammatically correct. This demonstrates that VIKDF can generate dialogues which are both contextually relevant and engaging, with a high degree of linguistic quality. The Fleiss' Kappa score, indicative of inter-rater agreement, falls within a moderate range across all models, ensuring that the evaluation process maintains a degree of reliability despite its inherently subjective nature.

In summary, these results affirm VIKDF's exemplary performance in zero-resource dialogue generation, particularly its adeptness at leveraging visual implicit knowledge to augment the relevance and informativeness of dialogues, while maintaining a high standard of fluency.

\begin{table}[t!]
  \caption{Ablation study. \textit{Zero Data} means in a fully zero-resource scenario.}
  \label{tab_ablation}
  \centering
  \resizebox{.45\textwidth}{!}{
  \begin{tabular}{l *{4}{c}}
    \toprule
    \multirow{2}{*}{Methods}
    & \multicolumn{2}{c}{Reddit Conversation} & \multicolumn{2}{c}{Image-Chat (\textit{Zero Data})} \\
    \cmidrule(lr){2-3} \cmidrule(lr){4-5}
    & {BLEU-1} & {ROUGE-L} & {BLEU-1} & {ROUGE-L}\\
    \midrule
    VIKDF & \textbf{16.41} & \textbf{14.82} & \textbf{15.35} & \textbf{15.30} \\
    \midrule
    -TIM & 14.75 & 13.51 & 13.80 & 13.96 \\
    -TAMIM & 16.03 & 14.18 & 14.81 & 15.01 \\
    -IAMTM & 16.18 & 14.18 & 15.11 & 15.11 \\
    -BVIF & 15.65 & 13.79 & 14.60 & 14.63 \\
    \bottomrule
  \end{tabular}
  }
\end{table}

\subsection{Ablation Study}
To evaluate the individual contributions of the components within VIKDF, we conduct an ablation study by sequentially removing each component and assessing the impact on performance metrics (BLEU-1 and ROUGE-L) across both Reddit Conversation and Image-Chat datasets. In the absence of BVIF, we adopt the vision-to-language generative learning as outlined in \cite{ref7}. The results, detailed in Table \ref{tab_ablation}, demonstrate a consistent decline in performance upon the exclusion of any component, underscoring their collective importance to the framework's efficacy. Notably, the exclusion of TIM yields the most pronounced decline in performance metrics. This indicates TIM's critical function in maintaining alignment between textual and visual modalities, a foundational aspect of generating coherent and contextually relevant dialogues. Additionally, omitting BVIF leads to a noticeable dip in performance metrics. This underscores BVIF's importance in seamlessly integrating distilled visual knowledge into the dialogue generation process, further enhancing the model's ability to produce contextually rich and engaging dialogues. The removal of TAMIM and IAMTM also leads to decreased performance. This result highlights their significance in enriching the model's capability to infer and align multimodal knowledge, thereby facilitating a more nuanced dialogue generation process. 

\begin{table}[t!]
  \caption{Performance comparison of VIKDF integrated with different LLMs.}
  \label{tab_llm_comparison}
  \centering
  \resizebox{.48\textwidth}{!}{
  \begin{tabular}{l *{4}{c}}
    \toprule
    \multirow{2}{*}{Methods}
    & \multicolumn{2}{c}{Reddit Conversation} & \multicolumn{2}{c}{Image-Chat (\textit{Zero Data})} \\
    \cmidrule(lr){2-3} \cmidrule(lr){4-5}
    & {BLEU-1} & {ROUGE-L} & {BLEU-1} & {ROUGE-L} \\
    \midrule
    Llama$_{LoRA}$        & 14.81 & 13.13 & 13.34 & 12.04 \\
    Qwen$_{LoRA}$         & 15.04 & 13.26 & 13.44 & 12.01 \\
    Mistral$_{LoRA}$      & 15.15 & 13.54 & 13.48 & 12.10 \\
    \midrule
    VIKDF$_{Llama}$       & 16.47 & 14.84 & 15.35 & 15.30 \\
    VIKDF$_{Qwen}$        & 16.27 & 14.65 & 15.22 & 15.29 \\
    VIKDF$_{Mistral}$     & 16.61 & 14.94 & 15.53 & 15.47 \\
    \bottomrule
  \end{tabular}
  }
\end{table}

Furthermore, to assess the generalizability and robustness of VIKDF across different large language models, we integrated our framework with three state-of-the-art open-source LLMs: Llama-2, Qwen2.5 \cite{ref49, ref50}, and Mistral-v0.3 \cite{ref51}, each with a parameter size of 7B. We first fine-tuned each LLM using LoRA \cite{ref52} to establish baseline performances. Subsequently, we applied VIKDF to each LLM, following the same knowledge distillation and integration processes described earlier. The performance metrics, presented in Table \ref{tab_llm_comparison}, reveal that integrating VIKDF consistently enhances the BLEU-1 and ROUGE-L scores across all models and datasets. Notably, the improvements are more pronounced in the zero-resource conditions on the Image-Chat dataset. This significant enhancement underlines VIKDF's effectiveness in leveraging implicit multimodal information to boost dialogue generation, especially when multimodal data is scarce. The consistent performance gains across different LLM architectures demonstrate the adaptability of our framework, showcasing its ability to seamlessly integrate with various models and improve their capabilities in challenging zero-resource scenarios. This underscores the robustness of VIKDF and its potential for broad applicability in advancing dialogue systems.

\begin{figure*}
	\centering
	\includegraphics[scale=0.5]{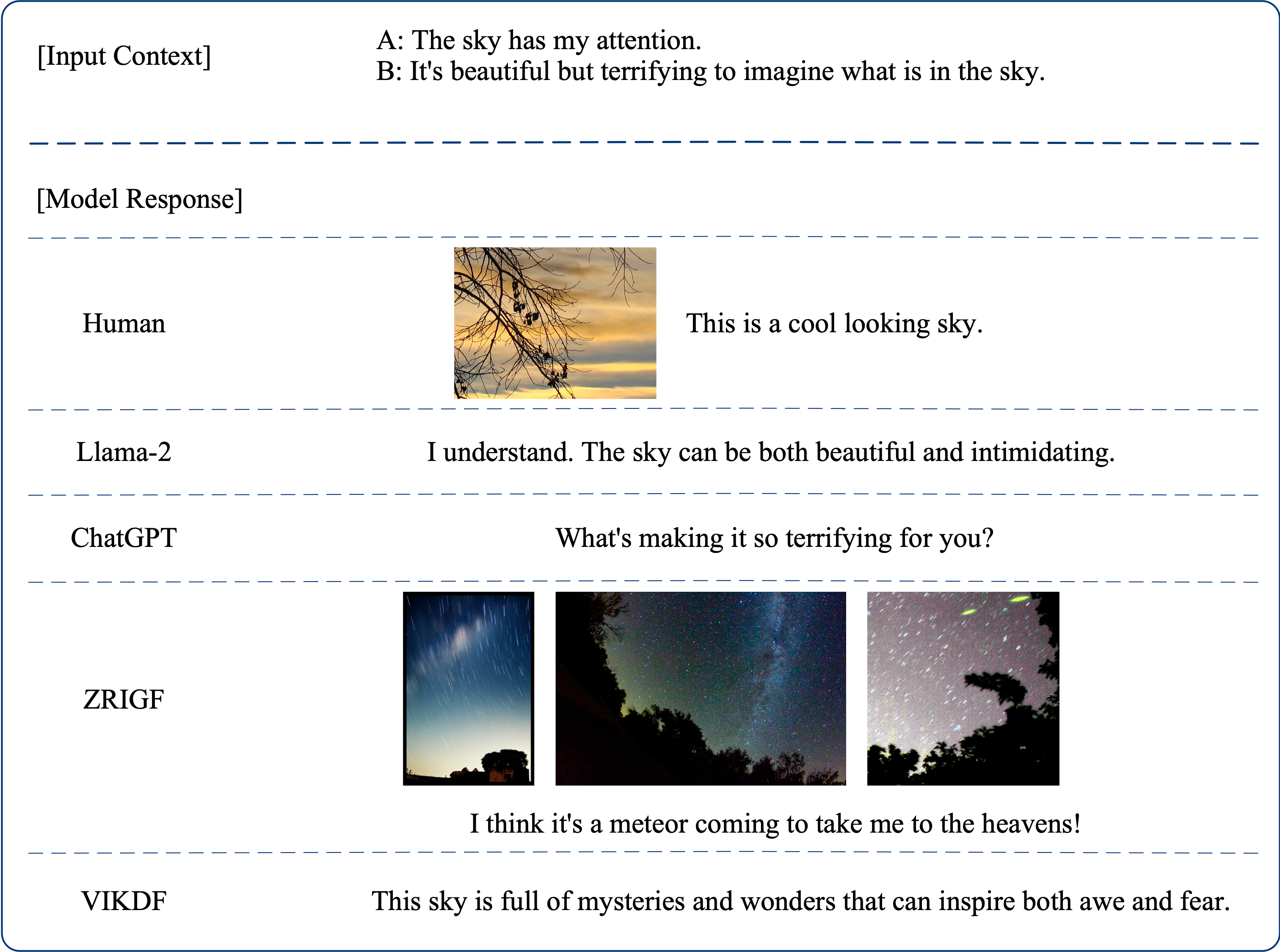}
	\caption{Case study on Image-Chat test set in a fully zero-resource scenario.}
	\label{fig_case}
\end{figure*}

\begin{figure}[ht!]
	\centering
	\includegraphics[scale=0.5]{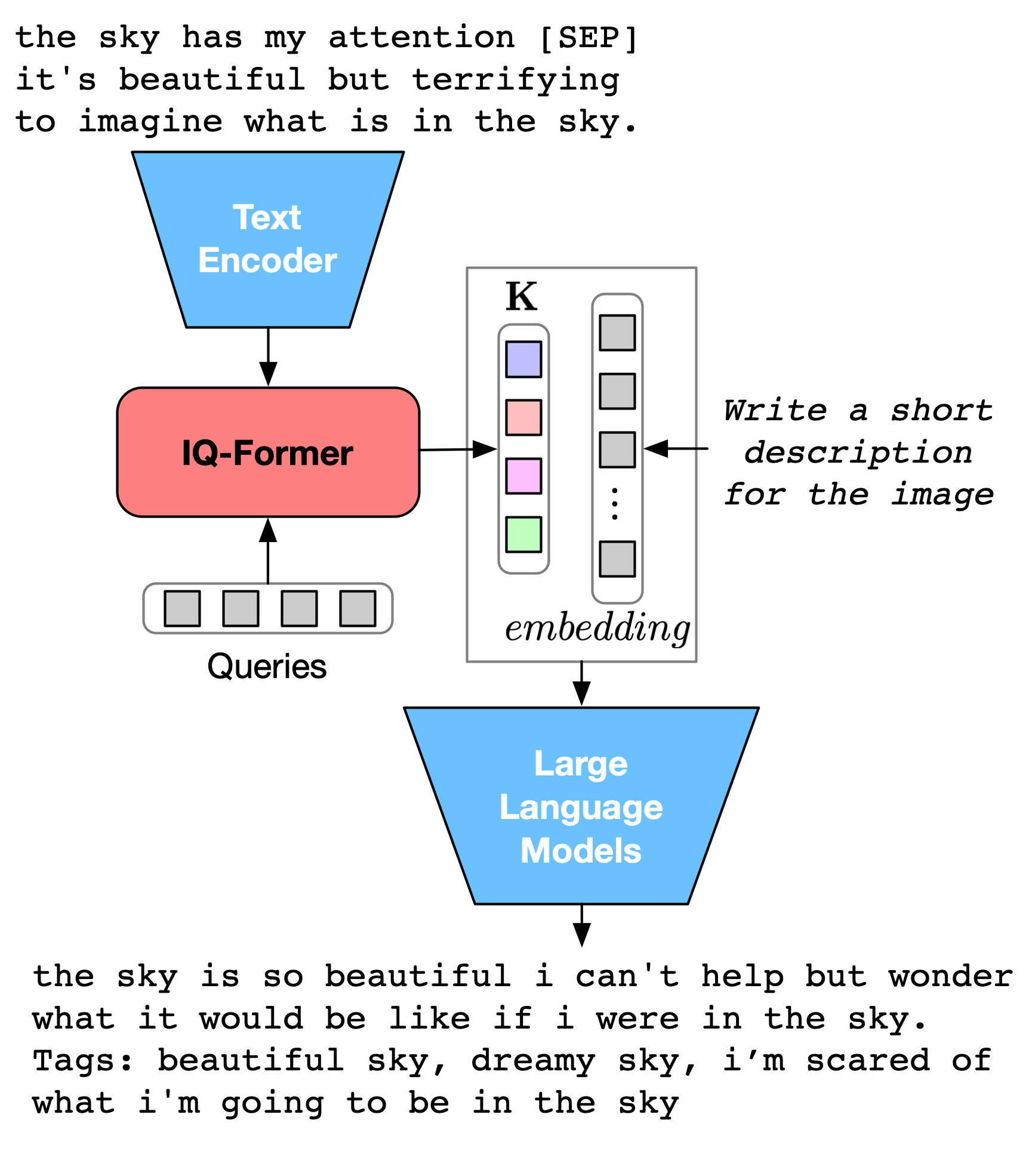}
	\caption{Example of visual implicit knowledge textualized from dialogue context by VIKDF.}
	\label{fig_infer}
\end{figure}

\subsection{Case study}
To further illustrate the superior capabilities of VIKDF, we conduct a case study contrasting VIKDF against key baseline models by examining a specific example from the Image-Chat test set in a fully zero-resource scenario, as shown in Figure \ref{fig_case}. In a dialogue context that evokes curiosity and apprehension towards the sky, VIKDF leverages distilled visual implicit knowledge to generate a response that captures both the awe of the sky's vast beauty and the curiosity towards its unknown mysteries. In comparison, Llama-2 produces a relevant but less informative response. ChatGPT, while producing a context-aware response, misses the mark on integrating the emotive complexity of the conversation, focusing instead on extracting additional context from the user. This demonstrates the limitations of LLMs in multimodal dialog generation scenarios. In contrast, ZRIGF's response, though creative, diverges from the contextual context due to its reliance on retrieved explicit images, whereas VIKDF overcomes this by distilling visual implicit knowledge from text.

To provide a more detailed analysis of visual implicit knowledge, we attempt to textualize the visual implicit knowledge through the LLM, following the process illustrated in Figure \ref{fig_infer}. The process involves transforming the dialogue context $C$ through the IQ-Former, which distills visual implicit knowledge $K$ solely from $C$, and subsequently textualizing $K$ via the LLM using a text prompt. Importantly, no image input is provided at this stage; the model relies entirely on the textual dialogue context. We can see that the textual output illustrates the model's capacity to encapsulate complex human emotions and curiosities about the sky, derived from the dialogue context. This demonstrates the model's adeptness at integrating and expressing visual implicit knowledge. While the LLM possesses substantial world knowledge due to pre-training, the integration of the distilled knowledge $K$ allows it to capture visual implicit knowledge that is contextually relevant and may not be explicitly encoded in the LLM's pre-trained weights. Thus, the LLM produces responses that reflect both its pre-existing knowledge and the newly distilled visual implicit knowledge pertinent to the dialogue context.

This analysis underscores VIKDF's superior ability to synthesize and leverage visual implicit knowledge, enabling it to generate dialogues that are more engaging, visually grounded, and contextually appropriate.

\begin{figure*}
	\centering
	\begin{subfigure}{0.45\textwidth}
	\includegraphics[width=\linewidth]{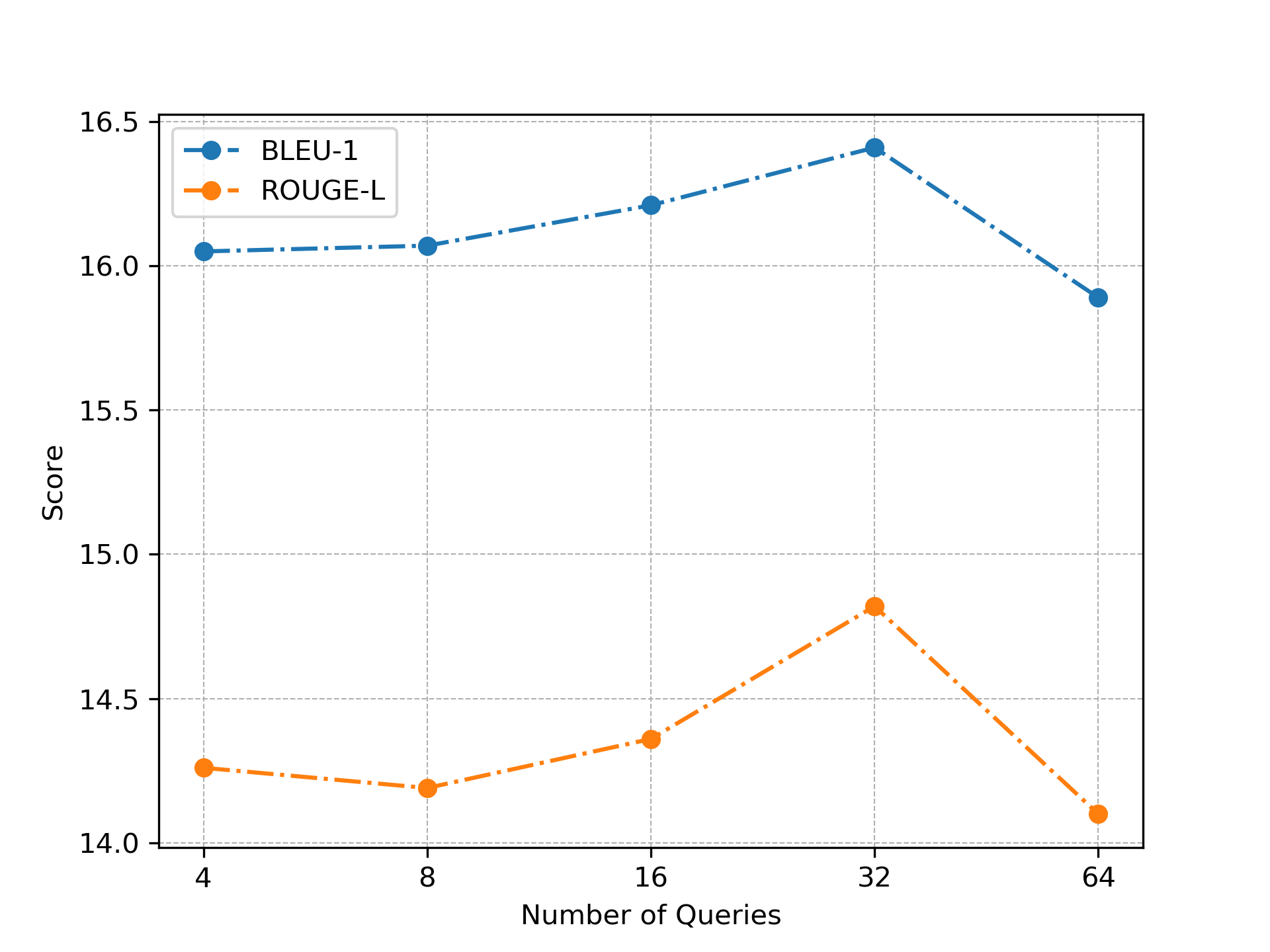}
	\caption{Reddit Conversation}
	\end{subfigure}%
	\begin{subfigure}{0.45\textwidth}
	\includegraphics[width=\linewidth]{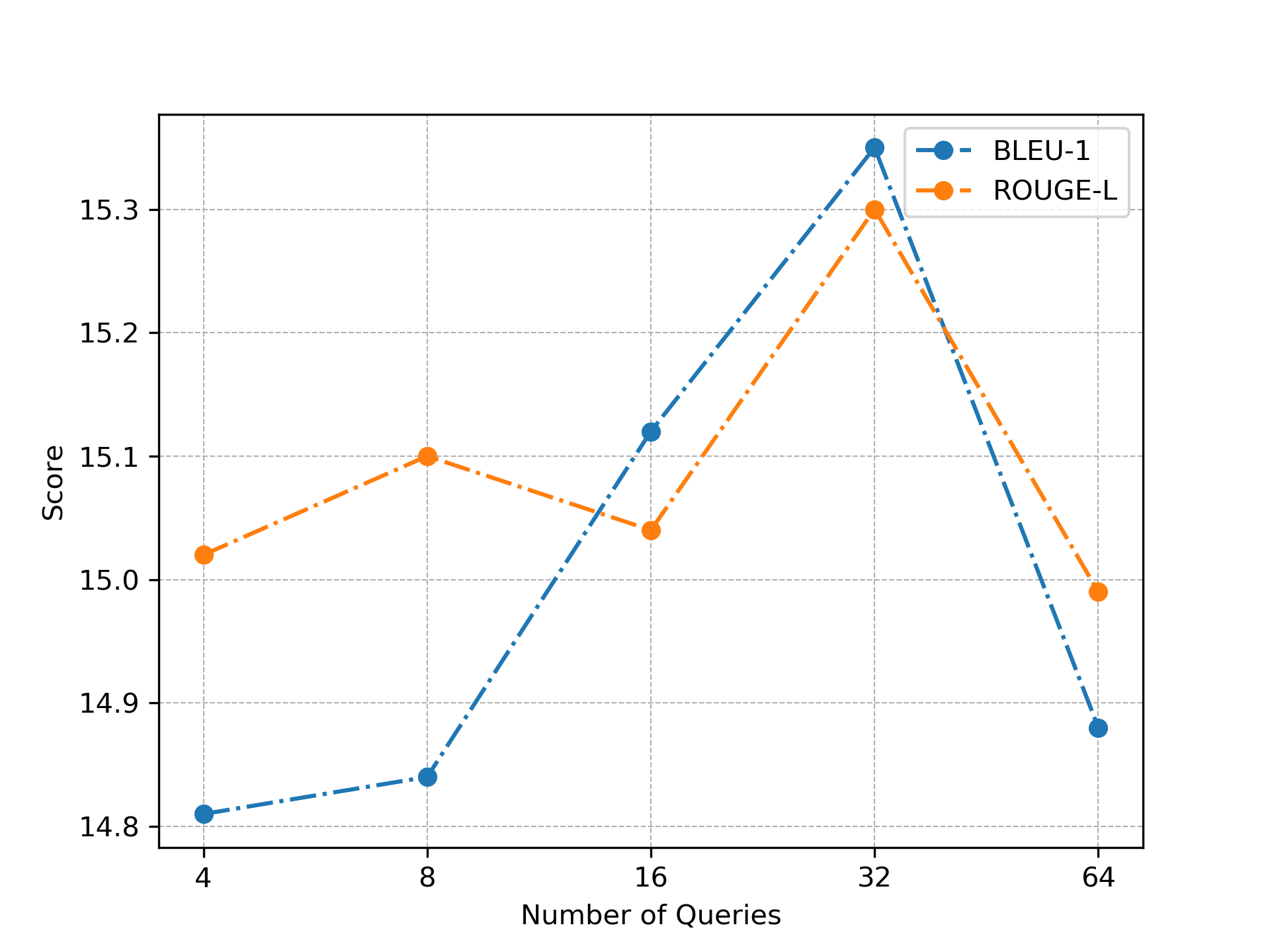}
	\caption{Image-Chat (\textit{Zero Data})}
	\end{subfigure}
	\caption{Performance comparison with various numbers of queries.}
	\label{fig_qnum}
\end{figure*}

\subsection{Impact of Query Vector Quantity}
Exploring the influence of the hyper-parameter $n$, which dictates the number of query vectors deployed in the IQ-Former, is crucial for understanding the dynamics of the VIKDF in dialogue generation tasks. To this end, we adjust $n$ across 4, 8, 16, 32, and 64 to examine its impact on model performance, focusing on BLEU-1 and ROUGE-L scores within the Reddit Conversation and Image-Chat datasets. Figure \ref{fig_qnum} presents the results, which reveal a nuanced relationship between the quantity of query vectors and model performance. Remarkably, a configuration of $n = 32$ is identified as optimal, yielding the highest BLEU and ROUGE scores across both datasets. This suggests an ideal balance in leveraging visual implicit knowledge: too few query vectors ($n < 32$) may not capture the breadth of implicit knowledge available, whereas too many ($n > 32$) could introduce unnecessary noise or dilute the relevance of the distilled knowledge. 

The experiment highlights the importance of a balanced query vector count in achieving effective dialogue generation. An optimal $n$ allows the IQ-Former to distill relevant visual implicit knowledge without overcomplicating the model, demonstrating the delicate balance between quantity and quality of distilled knowledge for enhancing dialogue generation.

\section{Conclusion and Future Work}
In this paper, we present VIKDF, an innovative methodology aimed at enhancing LLMs for dialogue generation in zero-resource scenarios through the distillation and integration of implicit multimodal knowledge. VIKDF utilizes an IQ-Former to extract visual implicit knowledge and a BVIF technique to incorporate this knowledge into LLMs, enabling the generation of dialogues that are coherent, engaging, and rich in contextual understanding. Our comprehensive experiments across diverse dialogue datasets have shown that VIKDF outperforms existing state-of-the-art models in zero-resource scenarios, illustrating its effectiveness in leveraging implicit multimodal knowledge even without explicit multimodal inputs or annotated datasets. The ablation study underscores the indispensable role of each component within VIKDF, and human evaluations have confirmed its success in generating dialogues that are relevant, informative, and naturally fluent, closely aligning human conversational standards. Consequently, VIKDF represents a significant advancement in the field of multimodal dialogue generation, highlighting the importance of implicit multimodal knowledge in enhancing LLMs capabilities in zero-resource scenarios.

The proposed model utilizes only implicit multimodal information, which limits its applicability in tasks requiring explicit multimodal inputs, such as visual question answering, and multimodal outputs, such as text-to-image generation. In future work, we plan to integrate both explicit and implicit multimodal information to develop a dialogue generation system capable of supporting both multimodal inputs and outputs. This advancement will enable our model to engage more comprehensively with various types of content, potentially enhancing its performance and applicability in multimodal interaction scenarios.

\section*{Acknowledgements}
This research is supported by the National Natural Science Foundation of China (No. 62006034) and the Anhui Provincial Natural Science Foundation (2408085QF188).



\bibliographystyle{elsarticle-num}
\bibliography{paper}


%\begin{thebibliography}{00}
%%% Text of bibliographic item
%
%\bibitem{}
%
%\end{thebibliography}
\end{document}